\documentclass{article}
\usepackage{lmodern}
\usepackage[latin9]{inputenc}
\usepackage{float}
\usepackage{amsthm}
\usepackage{amsmath}
\usepackage{graphicx}
\usepackage{esint}
\usepackage[numbers]{natbib}

\makeatletter

\newcommand{\noun}[1]{\textsc{#1}}
\floatstyle{ruled}
\newfloat{algorithm}{tbp}{loa}
\providecommand{\algorithmname}{Algorithm}
\floatname{algorithm}{\protect\algorithmname}

\usepackage[color=green!40]{todonotes}
\theoremstyle{plain}
\newtheorem{thm}{\protect\theoremname}
  \theoremstyle{definition}
  \newtheorem{defn}[thm]{\protect\definitionname}
 \newtheorem{hypothesis}{Hypothesis}

\usepackage{algorithmic}

\makeatother

  \providecommand{\definitionname}{Definition}
\providecommand{\theoremname}{Theorem}

\begin{document}

\title{Decentralized Multi-agent Plan Repair in Dynamic Environments\footnote{This is the full version of an extended abstract published in \emph{Proceedings of the 11th International Conference on Autonomous Agents and Multiagent Systems} (AAMAS 2012), Conitzer, Winikoff, Padgham, and van der Hoek (eds.), June, 4--8, 2012, Valencia, Spain.}}

\author{Anton\'{i}n Komenda, Peter Nov\'{a}k, Michal P\v{e}chou\v{c}ek\\
\\
Agent Technology Center\\
Dept.~of Computer Science and Engineering, FEE\\
Czech Technical University in Prague\\
Czech Republic}

\date{{}}
\maketitle
\begin{abstract}
Achieving joint objectives by teams of cooperative planning agents requires significant coordination and communication efforts. For a single-agent system facing a plan failure in a dynamic environment, arguably, attempts to repair the failed plan in general do not straightforwardly bring any benefit in terms of time complexity. However, in multi-agent settings the communication complexity might be of a much higher importance, possibly a high communication overhead might be even prohibitive in certain domains. We hypothesize that in decentralized systems, where coordination is enforced to achieve joint objectives, \emph{attempts to repair failed multi-agent plans should lead to lower communication overhead than replanning from scratch}.

The contribution of the presented paper is threefold. Firstly, we formally introduce the multi-agent plan repair problem and formally present the core hypothesis underlying our work. Secondly, we propose three algorithms for multi-agent plan repair reducing the problem to specialized instances of the multi-agent planning problem. Finally, we present results of experimental validation confirming the core hypothesis of the paper.
\end{abstract}

\section{Motivation \label{sec:Motivation}}

Classical planning and multi-agent planning based on classical planning are approaches to constructing autonomous agents and teams of agents, which attempt to achieve their objectives in an environment. The result of the planning process is traditionally a plan, a sequence of actions the agent should perform in order to achieve a given goal. When the agent is situated in a dynamic environment, occurrence of various unexpected events the environment generates might lead to the plan invalidation, a failure. A straightforward solution to this problem is to invoke a planning algorithm and compute a new plan from the state the agent found itself in after the failure to a state conforming with its original objective.

Planning, as well as replanning, in the case of a failure occurrence, is a costly procedure, especially in terms of its time complexity. It is relatively straightforward to see, that in many cases, however, a relatively minor fix to the original plan would resolve the failure possibly at a lower cost. Because it is not clear what exactly are the planning domains and types of dynamic environments which would allow for such a repair approach, it can be argued that non-informed plan repair attempts can in many cases even raise the overall complexity of the approach in comparison to replanning. This would be due to futile attempts to repair the failed plan before inevitably falling back to replanning. 

In general, plan repair can be seen as planning with re-use of fragments of the old plan. There is a number of works, empirically demonstrating that plan repair in various domains performs better than replanning (e.g., \citep{krogt05icapsws,Au2002,Fox2006}). However, in~\citep{NebelB1995}, Nebel and Koehler theoretically analyzed plan re-use (plan repair), and conclude that in general it does not bring any benefit over replanning in terms of computational time complexity.

In situated multi-agent systems, however, the time complexity is often not of the primary importance. In such systems, often it is the communication complexity which is a higher priority concern. Consider application domains, such as e.g., undersea operations by teams of coordinated autonomous underwater vehicles. While the state-of-the-art technology allows to employ relatively powerful computers on board of such robots, the communication links are extremely constrained and expensive; wireless networks cannot be deployed and communication is performed mostly using acoustic signaling. In such applications, it is the communication complexity of the distributed planning algorithms which matters more than time complexity. Consequently, employment of multi-agent plain repair techniques can provide a tangible benefit over replanning for a team of robots whose multi-agent plan fails. 

The motivation for our research is the intuition that multi-agent plan repair, even though not always the fastest approach, should under specific conditions generate lower communication overheads in comparison to replanning. The conditions correspond to the level of required coordination and the types of failures the environment generates. While the hypothesis is rather intuitive, our approach is significant in that we give it a rigorous treatment. Besides our preliminary approach in~\citep{conf/DMPOUW2011/KomendaNovak2011},\textbf{ }this aspect of multi-agent planning and plan repair, while obviously important with arrival of truly distributed algorithms for multi-agent planning, such as the one by Nissim et al.~\citep{DBLP:conf/atal/NissimBD10} did not witness considerable attention of the community yet. 

The contribution of the the presented paper is threefold. Firstly, after introducing the general problem of multi-agent planning stemming from the formulation due to Brafman and Domshlak~\citep{DBLP:conf/aips/BrafmanD08} in Section~\ref{sec:Multi-agent-planning}, in subsequent Section~\ref{sec:Multi-agent-plan-repair} we formally introduce the multi-agent plan repair problem and formally state the core hypothesis of the presented research. Secondly, still in Section~\ref{sec:Multi-agent-plan-repair}, we propose three algorithms for multi-agent plan repair reducing the problem to specialized instances of the multi-agent planning problem. Finally, in Section~\ref{sec:Evaluation} we present experimental validation confirming the core hypothesis of the paper. Section~\ref{sec:Discussion} concludes the paper by some final remarks regarding the shortcomings of our approach and future outlooks in the here described line of research.

\section{Multi-agent planning }

\label{sec:Multi-agent-planning}

\global\long\def\States{\mathcal{S}}

\global\long\def\Agents{\mathcal{A}}

\global\long\def\undef{\chi}

\global\long\def\Actions{\mathit{Act}}

\global\long\def\ja{\mathbf{a}}

\global\long\def\diff{\mathit{diff}}

\global\long\def\Plans{\mathit{Plans}}

\global\long\def\sfail{s_{\mathit{f}}}

Classical, single-agent planning problem is characterized by a set of states with a unique initial state, a final state (a set of final states) and a set of actions representing the transitions between these states that the system undertakes upon performing the actions. We define multi-agent planning problem as an extension of the classical single-agent planning. We consider a number of \emph{cooperative} and \emph{coordinated} actors featuring possibly distinct sets of capabilities (actions), which concurrently plan and subsequently execute their local plans so that they achieve a joint goal.

An instance of a multi-agent planning problem is defined by: i) an environment characterized by a state space, ii) a finite set of agents, each characterized by a set of primitive actions (or capabilities) it can execute in the environment, iii) an initial state the agents start their activities in and iv) a characterization of the desired goal states. Before treating the problem of multi-agent planning and summarizing a state-of-the-art algorithm for solving it, we first formally introduce the underlying concepts.

\subsection{Preliminaries}

Consider a set of atoms $\{p_{1},\ldots,p_{n}\}$. A \emph{state} is a set of terms from a language $\mathcal{L}=\{p_{1},\ldots,p_{n},\neg p_{1},\ldots,\neg p_{n}\}$ where $\neg p$ denotes a negation of $p$. We also assume the standard tautology $p\equiv\neg\neg p$ for every $p\in\mathcal{L}$. Furthermore, we require all states to be \emph{consistent}, i.e., for a state $s\subseteq\mathcal{L}$ we have that $p\in s$ if and only if $\neg p\notin s$. Note, in general, the states do not have to be complete, i.e., it might be that there is a $p\in\mathcal{L}$, such that $\{p,\neg p\}\not\subseteq s$. $\States$ denotes the set of all states and we assume there is a distinguished state $\undef\in\States$ denoting an undefined state in which the overall system can be in. To simplify the notation, we also extend the negation to states as follows $\neg s=\{p|\neg p\in s\}$. The set of atoms corresponding to a set of terms $\phi\subseteq\mathcal{L}$ is denoted $\underline{\phi}=\{p\mid p\in\phi\textrm{ or }\neg p\in\phi\}$.

A \emph{primitive} \emph{action}, or simply an \emph{action}, is a tuple $\phi a\psi$, where $a$ is a unique action label and $\phi,\psi\subseteq\mathcal{L}$ respectively denote the sets of preconditions and effects of $a$. The preconditions and effects are assumed to be consistent sets of terms. Whenever the context is clear, we simply write $a$ instead of $\phi a\psi$. $\Actions$ denotes the set of all actions and we furthermore assume there is a distinguished empty action $\emptyset\epsilon\emptyset\in\Actions$ with no preconditions and no effects. 

We say that an action $\phi a\psi$ is \emph{applicable} in a state $s$ iff $\phi\subseteq s$. An application of $\phi a\psi$ is defined by the state transformation operator $\oplus:\States\times\Actions\rightarrow\States$ defined as follows: 
\[
s\oplus a=\begin{cases}
(s\cup\psi)\setminus\neg\psi & \textrm{ iff }\phi\textrm{ is applicable in }s,\\
\chi & \textrm{ otherwise}.
\end{cases}
\]

Note that given a consistent state $s$ and an action $\phi a\psi$, the application of $a$ to $s$ results in either the undefined state $\undef$, or, in the case $a$ was applicable in $s$, a consistent state $s^{\prime}$ again. The application of $\phi a\psi$ to $s$ first enriches $s$ with all the effects of $a$, however, in the case there exists some $p\in s$, s.t., $\neg p\in\psi$, the simple unification would make the resulting state inconsistent. The subsequent set subtraction of terms which were the source of such inconsistencies makes the resulting state consistent again, while at the same time preserving the effects of $a$. Furthermore, $\oplus$ is associative, hence we can write $s\oplus a_{1}\oplus\cdots\oplus a_{n}$.

An \emph{agent} $\alpha=\{\phi_{a_{1}}a_{1}\psi_{a_{1}},\ldots,\phi_{a_{n}}a_{n}\psi_{a_{n}}\}$ is characterized precisely by its capabilities, a finite repertoire of actions it can preform in the environment. From now on, we assume that there exists a language $\mathcal{L}$ giving rise to a state space $\States$.

\subsection{The Problem of Multi-agent Planning}
\begin{defn}[multi-agent planning]
 \label{def: multi-agent-planning} A \emph{multi-agent} \emph{planning problem} is a tuple $\Pi=(\Agents,s_{0},S_{g})$, where $\Agents$ is a set of \emph{agents} $\alpha_{1},\ldots,\alpha_{n}$, featuring mutually disjoint sets of actions, an initial state $s_{0}\in\States$ and a set of goal states $S_{g}\subseteq\States$.
\end{defn}
Before formally defining the notion of a solution to a multi-agent planning problem, we first introduce a sequel of auxiliary notions.

Given an agent $\alpha$, a \emph{single-agent plan} $P$ is a sequence of actions $a_{1},\ldots,a_{k}$, s.t., $a_{i}\in\alpha$ for every $i$. $P[i]$ denotes the $i$-th action in $P$, or $P[i]=\epsilon$ in the case $i$ is larger than the length of $P$, which in turn will be denoted $|P|$.

A team of agents $\Agents=\alpha_{1},\ldots,\alpha_{n}$ can act in the environment concurrently. A joint action $\phi_{\ja}\ja\psi_{\ja}$ of the team is specified by $\ja=(\phi_{a_{1}}a_{1}\psi_{a_{1}},\ldots,\phi_{a_{n}}a_{n}\psi_{a_{n}})$ a tuple of actions corresponding to the individual agents, i.e., $a_{i}\in\alpha_{i}$ for each $i$, its preconditions $\phi_{\ja}=\bigcup_{i=1}^{n}\phi_{a_{i}}$ and its effects $\psi_{\ja}=\bigcup_{i=1}^{n}\psi_{a_{i}}$. $\ja[k]$ denotes the $k$-th action of $\ja$. Similarly to actions of individual agents, $\phi_{\ja}$ and $\psi_{\ja}$ are assumed to be consistent sets of terms. The notion of action applicability in a state $s$, as well as application of $\ja$ to $s$ straightforwardly extend from the definitions for primitive actions.
\begin{defn}[multi-agent plan]
 \label{def:multi-agent-plan} Let $\Pi=(\Agents,s_{0},S_{g})$ be a multi-agent planning problem with $\Agents=\alpha_{1},\ldots,\alpha_{n}$. A \emph{synchronous multi-agent plan} $\mathcal{P}=\{P_{1},\ldots,P_{n}\}$, consisting of single agent plans $P_{1},\ldots,P_{n}$ respectively constructed from actions of the agents $\alpha_{1},\ldots,\alpha_{n}$ is a solution to $\Pi$ if the plan $\mathcal{P}$ satisfies the following:
\begin{enumerate}
\item $\mathcal{P}$ is \emph{well-formed}, i.e., $|P_{i}|=|P_{j}|$ for all $i,j\leq n$. $|\mathcal{P}|=|P_{k}|$, for some $k\leq n$, denotes the length of the multi-agent plan $\mathcal{P}$;
\item $\mathcal{P}$ is \emph{feasible}, i.e., there exists a sequel of states $s_{1},\ldots,s_{m}$, s.t.~$m=|\mathcal{P}|$ and $s_{i+1}=s_{i}\oplus\ja_{i}$ with $\ja_{i}=(P_{1}[i],\ldots,P_{n}[i])$ for all $i<m$; and finally
\item $\mathcal{P}$ reaches the goal $S_{g}$, i.e., there exists $s_{g}\in S_{g}$, s.t.~$s_{m}\subseteq s_{g}$.
\end{enumerate}
We also say that $\mathcal{P}$ solves the problem $\Pi$. Finally, $\Plans(\Pi)$ denotes the set of plans which are solutions to a given multi-agent planning problem $\Pi$. 
\end{defn}
Additionally, $\mathcal{P}[k]$ denotes the joint action of the team in the step $k$ and $\mathcal{P}[k,i]$ denotes the primitive action of the agent $i$ in the step $k$. This notation allows us to introduce the following plan-matrix notation for a multi-agent plan $\mathcal{P}$, which provides a more visual understanding of multi-agent plans:

\[
\mathcal{P}=\left(\begin{array}{cccc}
a_{11} & a_{21} & \cdots & a_{m1}\\
a_{12} & a_{22} &  & a_{m2}\\
\vdots &  & \ddots & \vdots\\
a_{1n} & a_{2n} & \cdots & a_{mn}
\end{array}\right)
\]
where $a_{ij}=\mathcal{P}[i,j]$. Indices $i$ ($i\leq m=|\mathcal{P}|$) and $j$ ($j\leq n$) denote the step of the plan $\mathcal{P}$ and the agent which performs the primitive action, respectively.

We say that two multi-agent plans $\mathcal{P}_{1}$, $\mathcal{P}_{2}$ are equal ($\mathcal{P}_{1}=\mathcal{P}_{2}$) iff they have the same length ($|\mathcal{P}_{1}|=|\mathcal{P}_{2}|$) and for all $i$ and $j$ we have $\mathcal{P}_{1}[i,j]=\mathcal{P}_{2}[i,j]$.

A\emph{ }concatenation of two multi-agent plans $\mathcal{P}_{1}$ and $\mathcal{P}_{2}$ over the same agents $\alpha_{1},\ldots,\alpha_{n}$ is defined as a plan $\mathcal{P}=\mathcal{P}_{1}\cdot\mathcal{P}_{2}$, where for each $i$ and $j$ we have $\mathcal{P}[i,j]=\mathcal{P}_{1}[i,j]$ if $i\leq|\mathcal{P}_{1}|$ and $\mathcal{P}[i,j]=\mathcal{P}_{2}[i-|\mathcal{P}_{2}|,j]$ for $i>|\mathcal{P}_{1}|$. In the plan-matrix notation, the concatenation would correspond to simple columns-appending operation. Note, concatenation of multi-agent plans is an associative operation.

Given a multi-agent plan $\mathcal{P}$, $\mathcal{P}[i..j]$ denotes a fragment of $\mathcal{P}$ from the step $i$ to the step $j$. More precisely, $\mathcal{P}[i..j]$ is a fragment of $\mathcal{P}$ iff there exist multi-agent plans $\mathcal{P}_{\mathit{prefix}}$ and $\mathcal{P}_{\mathit{suffix}}$, such that $\mathcal{P}_{\mathit{prefix}}\cdot\mathcal{P}[i..j]\cdot\mathcal{P}_{\mathit{suffix}}=\mathcal{P}$. Finally, $\mathcal{P}[i..\infty]$ denotes the $i$-th suffix of the plan $\mathcal{P}$, i.e., $\mathcal{P}[i..\infty]=\mathcal{P}[i..|\mathcal{P}|]$. $\mathcal{P}_{1}\cdot\mathcal{P}_{2}$ is said to be a \emph{decomposition} of a multi-agent plan $\mathcal{P}$ iff $\mathcal{P}=\mathcal{P}_{1}\cdot\mathcal{P}_{2}$.

Given two multi-agent plans $\mathcal{P}_{1}$ and $\mathcal{P}_{2}$ we can define how different they are. $\diff(\mathcal{P}_{1},\mathcal{P}_{2})$ denotes the difference between $\mathcal{P}_{1}$ and $\mathcal{P}_{2}$, that is the overall number of primitive actions in $\mathcal{P}_{1}$, which do not correlate with the corresponding primitive actions in $\mathcal{P}_{2}$ and \emph{vice versa}. Formally, $\diff(\mathcal{P}_{1},\mathcal{P}_{2})=|\{(i,j)\mid\mathcal{P}_{1}[i,j]\neq\mathcal{P}_{2}[i,j]\}|$. In the case $|\mathcal{P}_{1}|\leq|\mathcal{P}_{2}|$, $\diff(\mathcal{P}_{1},\mathcal{P}_{2})=\diff(\mathcal{P}_{1}\cdot\mathcal{P}_{\epsilon},\mathcal{P}_{2})$, where $\mathcal{P}_{\epsilon}$ is a plan padding of $\mathcal{P}_{1}$ to the overall length $|\mathcal{P}_{1}\cdot\mathcal{P}_{\epsilon}|=|\mathcal{P}_{2}|$ and filled with empty actions, i.e, for each $i,j$, we have $\mathcal{P}_{\epsilon}[i,j]=\epsilon$. Note that the measure $\diff$ is position agnostic, i.e., we define $\diff(\mathcal{P}_{1},\mathcal{P}_{2})=\diff(\mathcal{P}_{2},\mathcal{P}_{1})$ in the case $|\mathcal{P}_{1}|>|\mathcal{P}_{2}|$.

\subsection{Planning Algorithm }

The above formulation of the multi-agent planning problem is well in line with the original formulation of \noun{MA-Strips} planning due to Brafman and Domshlak~\citep{DBLP:conf/aips/BrafmanD08}. The authors there additionally distinguish between the \emph{public} and \emph{private} actions of the individual agents. An action is public whenever its preconditions or effects involve atoms occurring in preconditions or effects of an action belonging to another agent of the team. Formally, given a multi-agent team ${\cal A}=\alpha_{1},\ldots,\alpha_{n}$, the set of public actions is defined as $\Actions^{\mathit{pub}}=\{a\mid\exists i,j:i\neq j,\, a\in\alpha_{i},\, a^{\prime}\in\alpha_{j},\textrm{ and }(\underline{\phi_{a}}\cup\underline{\psi_{a}})\cap(\underline{\phi_{a^{\prime}}}\cup\underline{\psi_{a^{\prime}}})\neq\emptyset\}$. Recall, $\underline{\phi}$ denotes the set of non-negated atoms occurring in $\phi$. $\Actions^{\mathit{priv}}=\Actions\setminus\Actions^{\mathit{pub}}$, where $\Actions=\bigcup_{i=1}^{n}\alpha_{i}$ is the set of all actions the team ${\cal A}$ can perform.

The distinction of actions to private and public turns out to be an important one. Since private actions do not depend, nor are dependencies of other actions performable by the team, planning of sequences of private actions can be implemented strictly locally by the agent the actions belongs to. In effect, the public actions become points of coordination among the multi-agent team members and a truly decentralized multi-agent planning algorithm for a planning problem $\Pi$ can be implemented in two interleaving stages until a suitable multi-agent plan is found: i) a plan consisting exclusively of public actions of the agent team is calculated, and subsequently ii) the sequences of private actions between the public actions of each individual agent are computed to fill in the gaps.

The main contribution of the Brafman and Domshlak's paper lies in pointing out that the algorithms can be implemented by reduction of the first stage of the planning process to a constraint satisfaction problem (CSP) corresponding to the multi-agent planning problem with public actions only. The second stage can be subsequently solved by any classical single-agent planning algorithm. In result, solving a given multi-agent planning problem can be loosely formulated as a CSP with the following two types of constraints:
\begin{description}
\item [{coordination~constraint:}] a sequence of joint actions $\mathcal{P}$ (candidate multi-agent plan) satisfies the \emph{coordination constraint} iff for every action $\phi_{a}a\psi_{a}=\mathcal{P}[k,i]$ performed by the agent $\alpha_{i}$ in the step $k$ we have, that if $a$ is a public action, then

\begin{itemize}
\item for every $p\in\phi_{a}$, there must exist $\phi_{a_{p}}a_{p}\psi_{a_{p}}=\mathcal{P}[k_{p},i_{p}]$, such that $p\in\psi_{a_{p}}$ and $0<k_{p}<k$ (there is some previous action which causes $p$ to hold), and
\item for no $k^{\prime}$, s.t.,~$k_{p}\leq k^{\prime}<k$ there exists $\phi^{\prime}a^{\prime}\psi^{\prime}=\mathcal{P}[k^{\prime},i^{\prime}]$, such that $\neg p\in\psi^{\prime}$ ($p$ won't be invalidated between causing it in the step $k_{p}$ and execution of $a$ in the step $k$).
\end{itemize}
\item [{internal~planning~constraint:}] a sequence of joint actions $\mathcal{P}$ satisfies the \emph{internal planning constraint} iff for every agent, the corresponding single-agent planning problem with landmarks $\{a\mid a=\mathcal{P}[k,i]\in\Actions^{\mathit{pub}}\}$ is solvable. I.e., a single-agent planning algorithm is able to fill in the gaps between the public actions in the candidate multi-agent plan.
\end{description}
\begin{algorithm}
\begin{algorithmic}
\REQUIRE A multi-agent planning problem $\Pi=(\Agents,s_0,S_g)$.
\ENSURE A multi-agent plan $\mathcal{P}$ solving $\Pi$, if such exists.
\medskip
\STATE $\delta=1$
\LOOP
\STATE Construct $\mathsf{CSP}_{\Pi;{\cal A}}$
\IF{solve-csp($\mathsf{CSP}_{\Pi;\delta}$)}
\STATE Reconstruct a plan $\mathcal{P}$ from a solution for $\mathsf{CSP}_{\Pi;\delta}$.
\RETURN $\mathcal{P}$
\ELSE
\STATE $\delta=\delta+1$
\ENDIF
\ENDLOOP
\end{algorithmic}

\caption{\label{alg:MA-Plan}\textsf{MA-Plan}($\Pi$):}
\end{algorithm}

Algorithm~\ref{alg:MA-Plan} lists the original multi-agent planning algorithm \textsf{MA-Plan} by Brafman and Domshlak in~\citep{DBLP:conf/aips/BrafmanD08}. The algorithm iterates through CSP formulations of the planning problem according to $\delta$, informally the number of coordination points between the agents in the multi-agent team. I.e., $\delta$ determines the number of joint actions in a candidate multi-agent plan consisting of only public actions. Filling the gaps between the individual single-agent public actions, if possible, then gives rise to the overall multi-agent plan. In the case such a plan completion does not exist, the process continues by testing longer candidate plans.

The original multi-agent planning algorithm assumes a centralized planning architecture. I.e., it is a centralized planning algorithm computing multi-agent plans for a team of agents which are supposed to be subsequently executed in a decentralized fashion. Our motivation is however a decentralized planning/plan repair algorithm followed by a decentralized plan execution.

Nissim et al.~in~\citep{DBLP:conf/atal/NissimBD10} adapted the original blueprint algorithm described above to a distributed setting. The adaptation rests on formulating the multi-agent planning problem as a distributed constraint satisfaction problem instance (\textsf{DisCSP}) and subsequently utilizing a a state-of-the-art \textsf{DisCSP} solver for solving it, plus managing the overhead involved in the resulting distributed algorithm. The resulting algorithm, however, closely follows the scheme of the original algorithm as listed in Algorithm~\ref{alg:MA-Plan}. From now on, whenever we speak about the implementation of the multi-agent planning algorithm, we have in mind its decentralized version due to Nissim et al.~\citep{DBLP:conf/atal/NissimBD10}.

\section{Multi-agent plan repair }

\label{sec:Multi-agent-plan-repair}

Consider a multi-agent planning problem $\Pi=(\Agents,s_{0},S_{g})$ and a plan $\mathcal{P}$ solving $\Pi$. Furthermore, consider an environment in which, apart from the actions performed by the agents of the team $\Agents$, no other exogenous events occur. We say that such an environment is \emph{ideal}, or \emph{non-dynamic}. The execution of $\mathcal{P}$ in such an environment is uniquely determined by the set of states $s_{0},\ldots,s_{m}$, such that $s_{i+1}=s_{i}\oplus\mathcal{P}[i]$ (cf.~also Definition~\ref{def:multi-agent-plan}). 

In dynamic environments, however, it can occur that in the course of execution of $\mathcal{P}$, the environment interferes and the execution of some action $\mathcal{P}[i]$ from the plan $\mathcal{P}$ does not result in precisely the state $s_{i+1}$ as defined above. We could say that at step $i$ an unexpected event occurred in the environment. For simplicity, we consider only unexpected events happening exclusively in the course of execution of some action (as if it took a non-zero time), not such which could occur while the agent is deliberating the execution (i.e., as if the deliberation was instantaneous).

Note that not all unexpected events in a dynamic environment necessarily lead to problems with execution of the plan $\mathcal{P}$. However, there are at least two cases of such events, which can be considered a \emph{plan execution failure.}

A \emph{weak failure} of execution of the plan $\mathcal{P}$ at step $i$ w.r.t.~the multi-agent planning problem $\Pi$ is such, when the state $\sfail$ resulting from an attempt to perform an action $\phi_{\ja}\ja\psi_{\ja}=\mathcal{P}[i]$ for some $i$ does not satisfy some of the postconditions of $\ja$, i.e., $\psi_{\ja}\not\subseteq\sfail$.

A \emph{strong failure} of execution of the plan $\mathcal{P}$ at step $i$ w.r.t.~the planning problem $\Pi$ occurs whenever the $i$-th action of $\mathcal{P}$ cannot be executed due to its inapplicability. I.e., the execution of the plan up to the step $i$ resulted in states $s_{0},s_{1}^{\prime}\ldots,s_{i}^{\prime}$, possibly with some weak failures occurring in the course of execution of the plan fragment and $\mathcal{P}[i]$ is not applicable in $s_{i}^{\prime}$.

The weak and the strong plan execution failures are, however, just two examples of a plan failure. There certainly are application domains in which weak failures can be tolerated as far as the goal state is reached after execution of the multi-agent plan. Alternatively, there might be domains in which other types of plan execution failures can occur, e.g.,~any change of the state not caused by the involved agents can be considered a failure as well. To account for the  range of various types failures, from now on, we only require that a plan execution monitoring process determines some plan execution failure at a step $i$ which results in some failed state~$\sfail$.
\begin{defn}[multi-agent plan repair]
 \label{def:multi-agent-plan-repair} Let $\Pi=(\Agents,s_{0},S_{g})$ be a multi-agent planning problem. A \emph{multi-agent} \emph{plan repair problem} is a tuple $\Sigma=(\Pi,\mathcal{P},\sfail,k)$, where $\mathcal{P}$ is a multi-agent plan solving the planning problem $\Pi$, $k$ is the step of $\mathcal{P}$ in which its execution failed and $\sfail\in\States$ is the corresponding failed state.

\emph{A solution to the plan repair problem} $\Sigma$ is a multi-agent plan $\mathcal{P}^{\prime}$, such that $\mathcal{P}^{\prime}$ is a solution to the planning problem $\Pi^{\prime}=(\Agents,\sfail,S_{g})$. We say that $\mathcal{P}^{\prime}$ \emph{repairs} $\mathcal{P}$ in $\sfail$. In the case $\Plans(\Pi^{\prime})=\emptyset$, we say that the plan is \emph{irreparable} given the failure occurring at the state~$\sfail$.

Given two multi-agent plans $\mathcal{P}_{1}$ and $\mathcal{P}_{2}$ both repairing a multi-agent plan $\mathcal{P}$ for a problem $\Pi$ in a state $\sfail$, we say that $\mathcal{P}_{1}$ is\emph{ preserving $\mathcal{P}$ more }than $\mathcal{P}_{2}$ iff $\diff(\mathcal{P}_{1},\mathcal{P})\leq\diff(\mathcal{P}_{2},\mathcal{P})$ and denote the relation by $\mathcal{P}_{1}\preceq\mathcal{P}_{2}$. The \emph{minimal repair of the multi-agent plan} $\mathcal{P}$ is such a plan $\mathcal{P}_{\min}\in\Plans(\Pi^{\prime})$, which is minimal w.r.t.~the mutual differences between the plans solving $\Pi^{\prime}$. I.e., $\mathcal{P}_{\min}\in\underset{\mathcal{P}^{\prime}\in\Plans(\Pi^{\prime})}{\arg\min}\diff(\mathcal{P},\mathcal{P^{\prime}})$.
\end{defn}
Note, there might be several distinct minimal repairs of a given multi-agent plan. 

In general, the multi-agent plan repair problem can be reduced to solving a modified multi-agent planning problem and thus gives rise to a straightforward plan repair algorithm based on \emph{replanning} in two steps: 1) construct the multi-agent replanning problem $\Pi^{\prime}$ as prescribed in Definition~\ref{def:multi-agent-plan-repair}, and subsequently 2) utilize the \textsf{MA-Plan} algorithm (Algorithm~\ref{alg:MA-Plan} ) to solve the problem $\Pi^{\prime}$.

The original motivation underlying this paper was the hypothesis that attempts to repair failed multi-agent plans lead to lower communication overhead than replanning. Clearly, not all planning problems could benefit from such a mechanism. Since we focus on multi-agent planning problems, which in a sense \emph{enforce} \emph{coordination} among the members of a multi-agent team, we firstly introduce the concept of $k$-coordinated multi-agent planning problems.
\begin{defn}[$k$-coordination]
 \label{def:k-tightness} We say that a multi-agent plan $\mathcal{P}$ is \emph{$k$-co\-or\-di\-na\-ted} iff each fragment $\mathcal{P}$ of length $k$ contains at least one joint action containing a public action. Formally, for every $\mathcal{P}^{\prime}$, s.t.~$\mathcal{P}=\mathcal{P}_{\mathit{prefix}}\cdot\mathcal{P}^{\prime}\cdot\mathcal{P}_{\mathit{suffix}}$ with $|\mathcal{P}^{\prime}|=k$, there exist $i$ and $j$ so that $\mathcal{P}^{\prime}[i,j]\in\Actions^{\mathit{pub}}$ .

We say that a multi-agent problem $\Pi$ is \emph{$k$-coordinated} iff all the plans $\mathcal{P}\in\Plans(\Pi)$ solving $\Pi$, which cannot be compressed are $k$-coordinated. A plan $\mathcal{P}\in\Plans(\Pi)$ can be \emph{compressed} iff it contains a fragment $\mathcal{P}^{\prime}$, s.t.~$\mathcal{P}=\mathcal{P}_{\mathit{prefix}}\cdot\mathcal{P}^{\prime}\cdot\mathcal{P}_{\mathit{suffix}}$ and $\mathcal{P}_{\mathit{prefix}}\cdot\mathcal{P}_{\mathit{suffix}}\in\Plans(\Pi)$.
\end{defn}
We informally say that multi-agent planning problems leading to plans containing coordination points (public actions) placed relatively frequently throughout the plans are tightly coordinated. More formally, a multi-agent planning problem $\Pi$ is \emph{tightly coordinated} if it is $k$-coordinated and $k$ is relatively low in comparison to the lengths of plans from $\Plans(\Pi)$. In the case $k$ is relatively high w.r.t.~the plan lengths, we say that the problem is \emph{loosely coordinated} and finally, if the plans do not involve public actions, i.e., coordination is not needed at all, we say the problem is \emph{uncoordinated}.

The core hypothesis of the paper can be then formulated as follows:
\begin{hypothesis}
\label{hyp:repair-lower-overhead} Multi-agent plan repair approaches producing more preserving repairs than replanning tend to generate \emph{lower communication overhead} for tightly coordinated multi-agent problems.
\end{hypothesis}
A crisper, though perhaps a more challenging version of the hypothesis would express the communication overhead in terms of the average communication complexity:
\begin{hypothesis}
\label{hyp:repair-lower-comm-complexity} When applied to tightly coordinated planning problems, multi-agent plan repair algorithms producing more preserving repairs than replanning feature a \emph{lower average communication complexity} than replanning.
\end{hypothesis}
In the remainder of this paper, we approach resolution of Hypothesis~\ref{hyp:repair-lower-overhead}. Treatment of Hypothesis~\ref{hyp:repair-lower-comm-complexity} is beyond the scope of this paper and is left for future work.

\subsection{Back-on-track Repair}

Unexpected event occurring in an environment can cause a failure in execution of a plan performed by some multi-agent team in that environment. The result is that the overall state of the system is not the one expected by the undisturbed plan execution at the particular time step. A straightforward idea to fix the problem is to utilize a multi-agent planner to produce a plan from the failed state to the originally expected state and subsequently follow the rest of the original multi-agent plan from the step in which the failure occurred. The following multi-agent plan repair approach, coined \emph{back-on-track} (BoT)\emph{ }repair, is inspired by this idea, in fact a slight generalization of it.
\begin{defn}[back-on-track repair]
 \label{def:back-on-track-repair}Let $\Sigma=(\Pi,\mathcal{P},\sfail,k)$ be a multi-agent plan repair problem and $\Pi^{\prime}=(\Delta,\sfail,S_{g})$ being the corresponding modified multi-agent replanning problem. 

We say that a plan $\mathcal{P}^{\prime}\in\Plans(\Pi^{\prime})$ is a \emph{back-on-track} \emph{repair} of $\mathcal{P}$ iff there is a decomposition of $\mathcal{P}^{\prime}$, such that $\mathcal{P}^{\prime}=\mathcal{P}^{\mathit{back}}\cdot\mathcal{P}[i..\infty]$ for some $i\leq|\mathcal{P}|$.

$\mathcal{P}^{\prime}=\mathcal{P}^{\mathit{back}}\cdot\mathcal{P}[i..\infty]$ is said to be a \emph{proper back-on-track} \emph{repair} iff $|\mathcal{P}[i..\infty]|>0$. I.e., $\mathcal{P}^{\prime}$ preserves some non-empty suffix of $\mathcal{P}$.
\end{defn}
Informally, the back-on-track approach tries to preserve a suffix of the original plan and prefix it with a newly computed plan $\mathcal{P}^{\mathit{back}}$ starting in $\sfail$ and leading to some state along the execution of $\mathcal{P}$ in the ideal environment. Note, that all plans from $\Plans(\Pi^{\prime})$ are back-on-track repairs of the original plan. The length of the preserved suffix of the original plan provides a handle on the repair quality ordering of the plans. The longer the preserved suffix, the more preserving the plan is. On the other hand, even when the plan repair problem $\Sigma$ is indeed solvable, there might not be any valid proper back-on-track repair of the original planning problem.

Algorithm~\ref{alg:MA-back-on-track-repair} realizes a multi-agent plan repair procedure according to the back-on-track plan repair principle. Since the \textsf{MA-Plan} algorithm searches for the shortest plan from the initial state to a goal state, the \textsf{Back-on-Track-Repair} computes plans which return back to the original one in the shortest possible way.

\begin{algorithm}
\begin{algorithmic}
\REQUIRE A multi-agent plan repair problem $\Sigma=(\Pi,\mathcal{P},\sfail,k)$, with $\Pi=(\Agents,s_0,S_g)$ and a sequence of states $s_0,\ldots,s_m$ execution of $\mathcal{P}$ generates in the ideal environment.
\ENSURE A multi-agent plan $\mathcal{P^\prime}$ solving $\Sigma$ if a solution exists.
\medskip
\STATE Construct  $\Pi^\mathit{back}=(\Agents,\sfail,\{s_0,\ldots,s_{m}\})$
\IF{$\textsf{MA-Plan}(\Pi^\mathit{back})$ returns a solution $\mathcal{P}^\mathit{back}$}
\STATE Retrieve the state $s_j$ of $\mathcal{P}$ to which $\mathcal{P}^\mathit{back}$ returns
\RETURN $\mathcal{P}^\prime=\mathcal{P}^\mathit{back}\cdot\mathcal{P}[j\ldots\infty]$
\ENDIF
\end{algorithmic}

\caption{\label{alg:MA-back-on-track-repair}\textsf{ Back-on-Track-Repair($\Sigma$)}}
\end{algorithm}

\subsection{Simple Lazy Repair}

The back-on-track multi-agent plan repair approach seeks to compute a new prefix to some suffix of the original plan and repair the failure by their concatenation. An alternative approach, coined \emph{lazy}, attempts to preserve the remainder of the original multi-agent plan and close the gap between the state resulting from the failed plan execution and a goal state of the original planning problem.

Let $\sfail$ be the state resulting from a failure in execution of a multi-agent plan $\mathcal{P}$ in a step $k$. We say that a sequence of joint actions $\mathcal{P}^{\prime}$ is an \emph{executable remainder} of $\mathcal{P}$ from the step $k$ and the state $\sfail$ iff there exists a sequence of states $s_{k},\ldots,s_{|\mathcal{P}|}$, such that $s_{k}=\sfail$, $s_{i+1}=s_{i}\oplus\mathcal{P}^{\prime}[i-k+1]$ and for every step $i$ and every agent $j$, we have that $\mathcal{P}^{\prime}[i-k+1,j]=\mathcal{P}[i,j]$ in the case $\mathcal{P}[i,j]$ is applicable in the state $s_{i}$ and $\mathcal{P}^{\prime}[i-k+1,j]=\epsilon$ otherwise.

The following definition provides a formal definition of the lazy approach.
\begin{defn}[simple lazy repair]
 \label{def:lazy-repair} Let $\Sigma=(\Pi,\mathcal{P},\sfail,k)$ be a multi-agent plan repair problem and $\Pi^{\prime}=(\Agents,\sfail,S_{g})$ being the corresponding modified multi-agent replanning problem.

We say that a plan $\mathcal{P}^{\prime}\in\Plans(\Pi^{\prime})$ is a \emph{lazy} \emph{repair} of $\mathcal{P}$ iff there is a decomposition of $\mathcal{P}^{\prime}$, such that $\mathcal{P}^{\prime}=\mathcal{P}_{[k..\infty]}\cdot\mathcal{P}^{\mathit{lazy}}$, where $\mathcal{P}_{[k..\infty]}$ is the executable remainder of $\mathcal{P}$ from the step $k$, execution of which results in the state $s_{\mathit{lazy}}$ when starting in $\sfail$, and $\mathcal{P}^{\mathit{lazy}}$ is a solution to the multi-agent planning problem $\Pi^{\mathit{lazy}}=(\Agents,s_{\mathit{lazy}},S_{g})$.
\end{defn}
Algorithm~\ref{alg:MA-lazy-repair} realizes multi-agent plan repair based on the lazy repair approach described above.

\begin{algorithm}
\begin{algorithmic}
\REQUIRE A multi-agent plan repair problem $\Sigma=(\Pi,\mathcal{P},\sfail,k)$, with $\Pi=(\Agents,s_0,S_g)$ and $|\Agents|=n$.
\ENSURE A multi-agent plan $\mathcal{P^\prime}$ solving the problem $\Sigma$ according to the lazy approach, if a solution exists.
\medskip
\STATE Construct $\mathcal{P}_{[k..\infty]}$, the executable remainder of $\mathcal{P}$ from the step $k$ and state $\sfail$
\STATE Construct $\Pi^\mathit{lazy}=({\cal A},s_{|\mathcal{P}|},S_g)$
\STATE let $\mathcal{P}^\mathit{lazy}$ be a solution to $\textsf{MA-Plan}(\Pi^\mathit{lazy})$ if such exists.
\RETURN $\mathcal{P}_{[k..\infty]}\cdot\mathcal{P}^\mathit{lazy}$
\end{algorithmic}

\caption{\label{alg:MA-lazy-repair}\textsf{ Lazy-Repair($\Sigma$)}}
\end{algorithm}

The back-on-track approach always succeeds to compute some multi-agent plan repairing the original plan from the failed state in the case the replanning form scratch would compute such a plan from that state as well. The lazy approach is in this sense incomplete, as it might happen that the execution of the executable remainder of the original plan diverges to a state from which no plan to some goal state exists. Given that dynamic environment in general could generate irreparable failures, this incompleteness cannot be considered a shortcoming of the lazy approach in general. Of course in domains in which no irreparable unexpected event might occur, while at the same time the agents are allowed to perform actions potentially having irreversible and potentially harmful effects, the lazy approach has to be employed with caution.

\subsection{Repeated Lazy Repair}

In dynamic environment plan failures occur repeatedly, i.e., even after a repair of a failed plan, it is possible for the repaired plan to fail again. In this situation both the back-on-track, as well as the lazy multi-agent plan repair algorithms lead to prolonging the really executed plan. In the case of the back-on-track approach, this is inevitable, since upon the repair, the subsequent plan execution process immediately processes the newly added plan fragment. In the case of the lazy repair, however, upon occurrence of another failure during execution of the repaired plan, it is not always necessary to prolong the overall multi-agent plan. 

The intuition behind the \emph{repeated lazy plan repair} approach is that a failure during execution of an already repaired plan makes the previous repair attempt irrelevant and its result can be discarded, unless the failure occurred already in the plan fragment appended by the previous repair. The following definition formally introduces the extension of the lazy multi-agent plan repair approach. For clarity, from now on, we refer to the lazy multi-agent plan repair approach introduced in Definition~\ref{def:lazy-repair} as \emph{simple lazy repair}.
\begin{defn}[Repeated lazy repair]
 \label{def:repeated-lazy-repair}Let $\Pi=(\Agents,s_{0},S_{g})$ be a multi-agent planning problem with a solution $\mathcal{P}$ and $\Sigma_{1}=(\Pi,\mathcal{P},s_{\mathit{f}_{1}},k_{1})$ be a multi-agent plan repair problem with a lazy repair solution $\mathcal{P}_{\Sigma_{1}}=\mathcal{P}_{[k_{1}..\infty]}\cdot\mathcal{P}_{\mathit{suffix}}$ and $\Sigma_{2}=(\Pi,\mathcal{P}_{\Sigma_{1}},s_{\mathit{f}_{2}},k_{2})$ be a multi-agent plan repair problem the system is currently facing.

We say that $\mathcal{P}^{\prime}$ is a \emph{repeated lazy repair} of $\mathcal{P}_{1}$ iff
\begin{enumerate}
\item $\mathcal{P}^{\prime}$ is a simple lazy repair solution to $\Sigma^{\prime}=(\Pi,\mathcal{P},s_{\mathit{f}_{2}},k_{2})$ in the case $k_{2}\leq|\mathcal{P}_{[k_{1}..\infty]}|$; and
\item $\mathcal{P}^{\prime}$ is a simple lazy repair solution to $\Sigma_{2}$ otherwise.
\end{enumerate}
\end{defn}
The repeated lazy repair leads to a straightforward extension of the lazy plan repair algorithm listed in Algorithm~\ref{alg:MA-lazy-repair}. Note, that the repeated lazy repair algorithms enables a plan execution model which preserves significantly longer fragments of the original plan. That is, upon a failure, instead of trying to repair the failed plan right away, as both the back-on-track and simple lazy plan repair algorithms do, the system can simply proceed with execution of the remainder of the original plan and only after its complete execution the lazy plan repair is triggered. The approach simply ignores the plan failures during execution and postpones the repair the very end of the process, hence the ``\emph{lazy}'' label for the two algorithms.

\begin{algorithm}
\begin{algorithmic}
\REQUIRE A multi-agent planning problem $\Pi=(\Agents,s_0,S_g)$, a multi-agent plan $\mathcal{P}=\mathcal{P}_{[k..\infty]}\cdot\mathcal{P}_\mathit{suffix}$ solving $\Pi$, and two multi-agent plan repair problems $\Sigma_1=(\Pi,\mathcal{P},s_{\mathit{f}_1},k_1)$ with a solution $\mathcal{P}_{\Sigma_1}$ and $\Sigma_2=(\Pi,\mathcal{P}_{\Sigma_1},s_{\mathit{f}_2},k_2)$.
\ENSURE A multi-agent plan solving $\Sigma_2$ if a solution exists.
\medskip
\IF{$k_2 \leq |\mathcal{P}_{[k..\infty]}|$}
	\STATE Construct  $\Sigma^\prime=(\Pi,\mathcal{P},s_{\mathit{f}_{2}},k_{2})$
\ELSE
	\STATE $\Sigma^\prime=\Sigma_2$
\ENDIF
\RETURN a solution to \textsf{Lazy-Repair($\Sigma^\prime$)} if such exists
\end{algorithmic}

\caption{\label{alg:MA-repeated-lazy-repair}\textsf{ Repeated-Lazy-Repair($\Sigma_{1},\mathcal{P}_{\Sigma_{1}},\Sigma_{2}$)}}
\end{algorithm}

\section{Experimental validation}

\label{sec:Evaluation}

To verify the Hypothesis~\ref{hyp:repair-lower-overhead}, we conducted a series of experiments with implementations of the multi-agent plan repair algorithms described in Section~\ref{sec:Multi-agent-plan-repair}. Below, we firstly describe the experimental setup used for the experiments and subsequently interpret the data collected and revisit Hypothesis~\ref{hyp:repair-lower-overhead}.

\subsection{Experimental Setup}

The experiments were based on a two-stage algorithm. In the first phase, for a given domain a multi-agent plan was computed using the \textsf{MA-Plan} algorithm based on a distributed constraint satisfaction solver for computing the candidate coordination plans and implementation of a best-first-search action planning algorithm as part of \textsf{FF}~\citep{hoffmann:nebel:jair-01} for computing the local, single-agent plans. We used the implementation of the distributed multi-agent planner authored by Nissim et al.~also used for the experiments conducted in their paper~\citep{DBLP:conf/atal/NissimBD10}. In the second phase, we executed the multi-agent plan. In the course of the plan execution, we simulated the environment dynamics by producing various plan failures according to a variable failure probability. The plan execution was monitored and upon a failure detection a plan repair algorithm was invoked. Algorithm~\ref{alg:MA-monitor-exec} lists the pseudo-code of the process.

Before execution of each plan step, the joint action is checked for applicability in the current state. In the case it is not applicable, a plan repair algorithm is invoked and the execution continues on the repaired plan. Otherwise, the state is updated with the joint action.

The \textsf{execute-fail} function in the algorithm either updates the current state by the joint action provided as a parameter as if in the ideal environment, or to generate an unexpected event occurring in the simulated dynamic environment.

We distinguish two types of plan failures: \emph{action failures} and \emph{state perturbations}. Both failure types are parametrized by a uniformly distributed probability $P$, which determines whether a simulation step fails, or not. Both failure types are weak failures. That is, they are not handled immediately, but can preclude the plan execution and later result in a strong failure. Upon detection, a strong failure is handled by one of the plan repairing algorithms.

\begin{algorithm}[t]
\begin{algorithmic}

\REQUIRE An initial multi-agent planning problem $\Pi=(\Agents,s_0,S_g)$.

\medskip{}

\STATE $\mathcal{P}=\textsf{MA-Plan}(\Pi)$

\STATE $s=s_{0}$; $\mathit{step}=1$

\REPEAT

\IF{$\phi_{\mathcal{P}[\mathit{step}]} \not \subseteq s$} 

\STATE $\mathcal{P}=\textsf{Repair}(\Pi,\mathcal{P},s,\mathit{step})$

\STATE $\mathit{step}=1$

\ENDIF

\STATE $s=\textsf{execute-fail}(s,\mathcal{P}[\mathit{step}])$

\STATE $\mathit{step}=\mathit{step}+1$

\UNTIL{$\mathit{step} > |\mathcal{P}|$}

\end{algorithmic}

\caption{\label{alg:MA-monitor-exec} Plan execution and monitoring algorithm.}
\end{algorithm}

An \emph{action failure }is simulated by not-executing some of the individual agent actions from the actual plan step. The individual action is chosen according to a uniformly distributed probability. The individual action is removed from the joint action and the current state is updated by the modified joint action.

The other simulated failure type, \emph{state perturbation, }is pa\-ra\-me\-tri\-zed by a positive non-zero integer $c$, which determines the number of state terms, which are removed from the current state, as well as the number of terms which are added to it. The terms to be added or removed are selected also randomly from the domain language according to a uniform distribution.

The experimental setup was implemented as a centralized simulator of the environment integrating a decentralized multi-agent domain-independent planner \textsf{MA-Plan}. The individual agents are initialized by a planning domain, together with a particular planning problem instance. Each agent runs in its own thread and they deliberate asynchronously. The agents send peer-to-peer messages among themselves. Message passing is mediated by the centralized simulator as well. The messages are sent in the \textsf{DisCSP} phase by the integrated solver, which is a part of the \textsf{MA-Plan} planner.

The experiments were performed on \emph{Phenom Quad Core 9950} processor at 2.6GHz with \emph{Java Virtual Machine} limited to 2.5GB of RAM. The individual measurements were parametrized by the plan failure probability $P$ and each problem instance was executed 6--10 with various value samples. The resulting data are, in the figures, presented with natural distribution. The candlestick charts depict the difference between the minimal and the maximal measurements, together with the standard deviation.

\subsection{Test Problems, Algorithms and Metrics}

The experiments were conducted on three planning domains. The domains originate in the standard benchmark single-agent ICP planning domains published at~\citep{ipc}. Similarly to~\citep{DBLP:conf/atal/NissimBD10}, we chose domains, which are straightforwardly modifiable to multi-agent planning problems: \textsc{logistics} (3 agents),\textsc{ rovers} (3 agents), and\textsc{ satellites} (2--5 agents). 

The \textsc{logistics} domain is a tightly coordinated in that it requires relatively frequent coordination among the involved agents: airplanes and trucks need to wait for each other to load or unload the transported packages. The\textsc{ rovers} domain is loosely coordinated in that it requires coordination only at the end of plans: there is a single shared communication channel between one of the rovers and the receiving station. Finally, the\textsc{ satellites} domain is uncoordinated in that it does not need any coordination between the satellites acquiring images individually.

To evaluate validity of Hypothesis~\ref{hyp:repair-lower-overhead}, the multi-agent planning problems were tested on the experimental setup against a plan repair algorithm implementing replanning from scratch and two of the repair algorithms \textsf{Back-on-Track-Repair} (Algorithm~\ref{alg:MA-back-on-track-repair}) and \textsf{Repeated-Lazy-Repair} (Algorithm~\ref{alg:MA-repeated-lazy-repair}) introduced in Section~\ref{sec:Multi-agent-plan-repair}.

Efficiency problems of the \textsf{MA-Plan} implementation limited the experiments to plans with maximally two landmarks (coordination points). The measurements of \textsf{Back-on-Track-Repair} algorithm runs were negatively influenced by sensitivity of the planner implementation to the number of terms in the goal state. Additionally, the \textsf{Back-on-Track-Repair} algorithm could not leverage disjunctive goal form (cf.~Definition~\ref{def:back-on-track-repair}) and this was emulated by iterative process testing all term conjunctions in a sequence and thus resulting in multiple runs of the \textsf{DisCSP} solver instead of a single run with disjunctive goal.

We used three metrics to evaluate the measurements:
\begin{description}
\item [{execution~length}] is the overall number of joint actions the experimental setup executed. 
\item [{planning~time}] is the measured cumulative time consumed by the underlying \textsf{MA-Plan} planner used for generating initial and repairing plans; and finally
\item [{communication}] is measuring the number of messages passed between the agents during the planning or plan repair process. That is messages generated by the \textsf{DisCSP} solver in the \textsf{MA-Plan} planner.
\end{description}

\subsection{Results and Discussion}

\begin{figure}
\begin{centering}
\includegraphics[width=0.8\columnwidth]{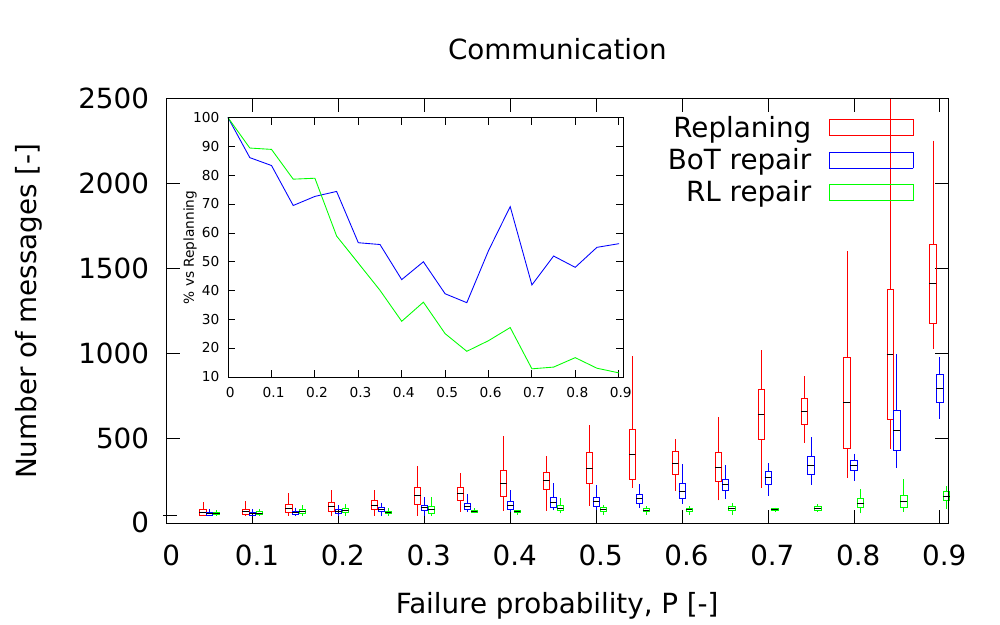}
\par\end{centering}

\begin{centering}
\includegraphics[width=0.8\columnwidth]{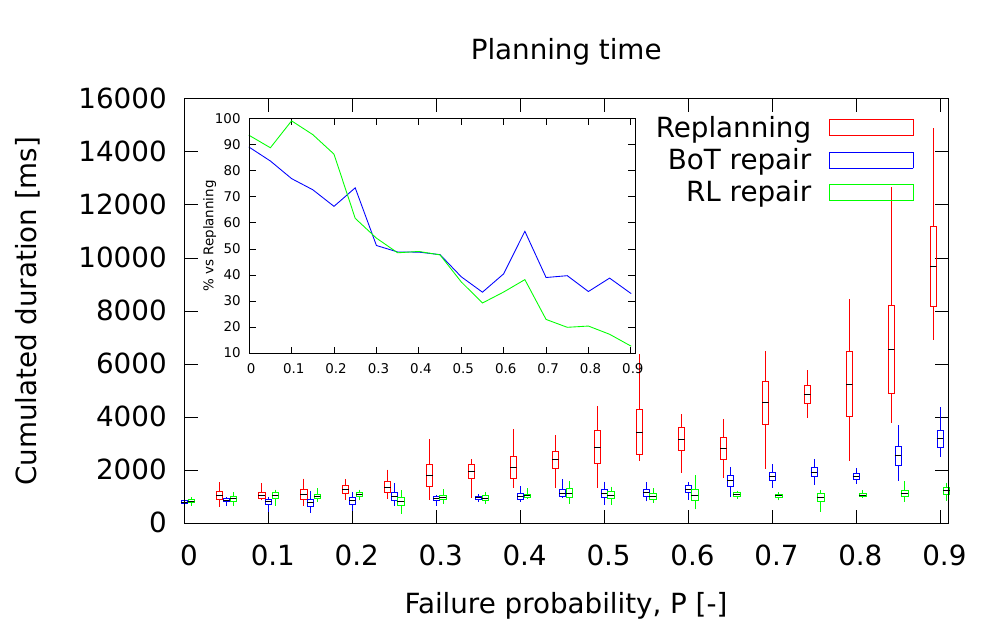}
\par\end{centering}

\begin{centering}
\includegraphics[width=0.8\columnwidth]{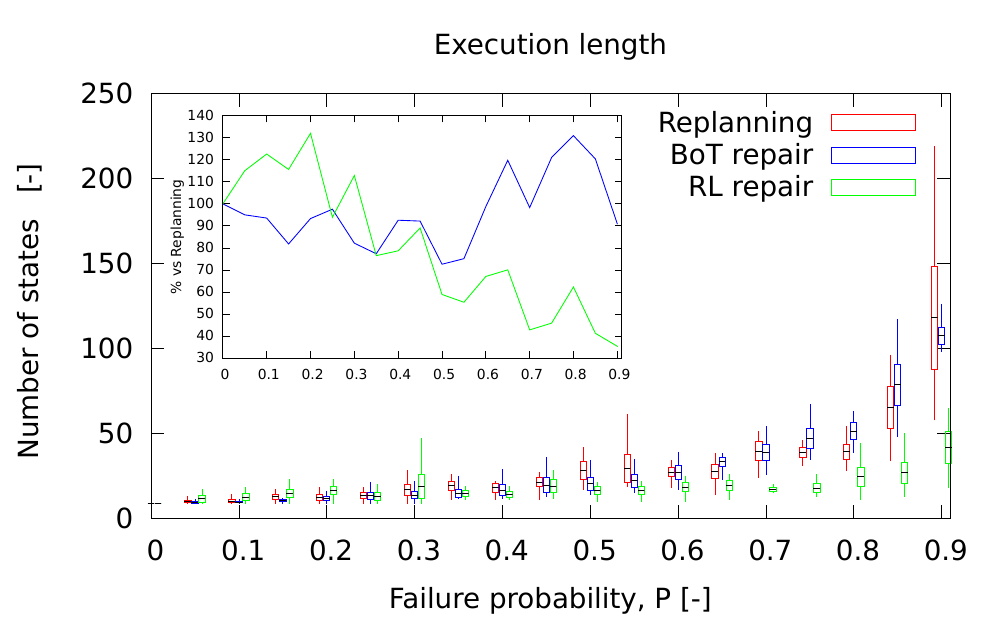}
\par\end{centering}

\caption{\label{fig:Experimental-results-for-1}Experimental results for \textsc{logistics} domain with 3 agents and action failures.}
\end{figure}

\begin{figure}
\begin{centering}
\includegraphics[width=0.8\columnwidth]{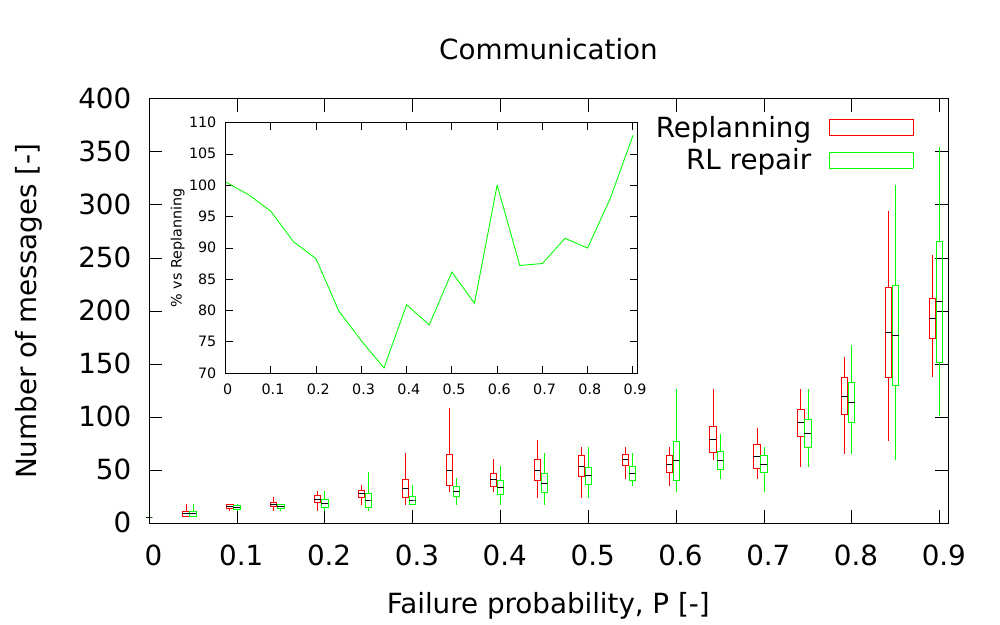}
\par\end{centering}

\begin{centering}
\includegraphics[width=0.8\columnwidth]{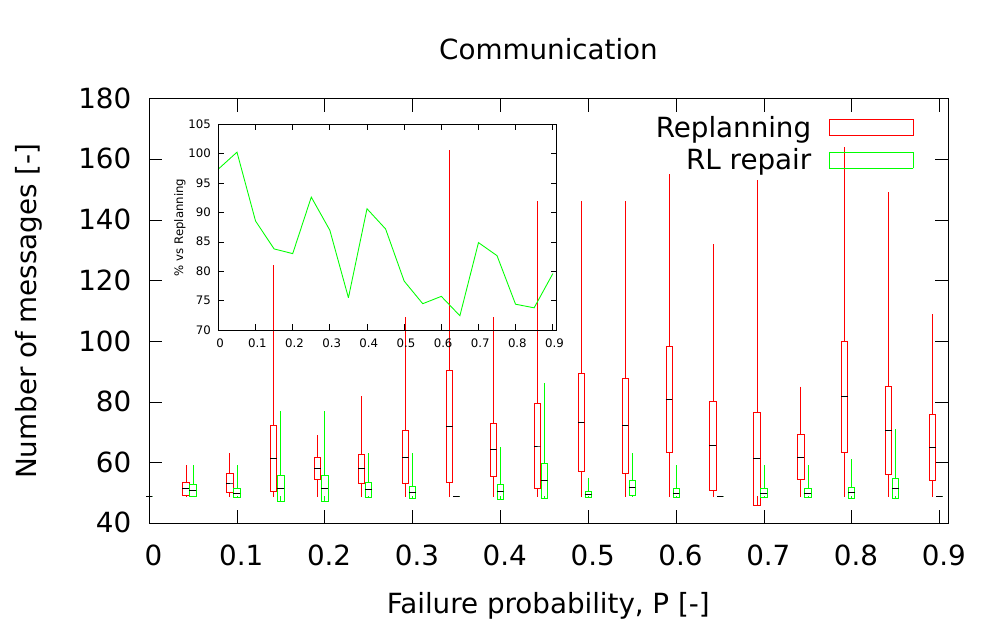}
\par\end{centering}

\caption{\label{fig:Experimental-log-sp}Experimental results for \textsc{rovers} domain with 3 agents and action failures (top). Experimental results for \textsc{logistics} domain with 3 agents and state perturbations with $c=1$ (bottom).}
\end{figure}

\begin{figure}
\begin{centering}
\includegraphics[width=0.8\columnwidth]{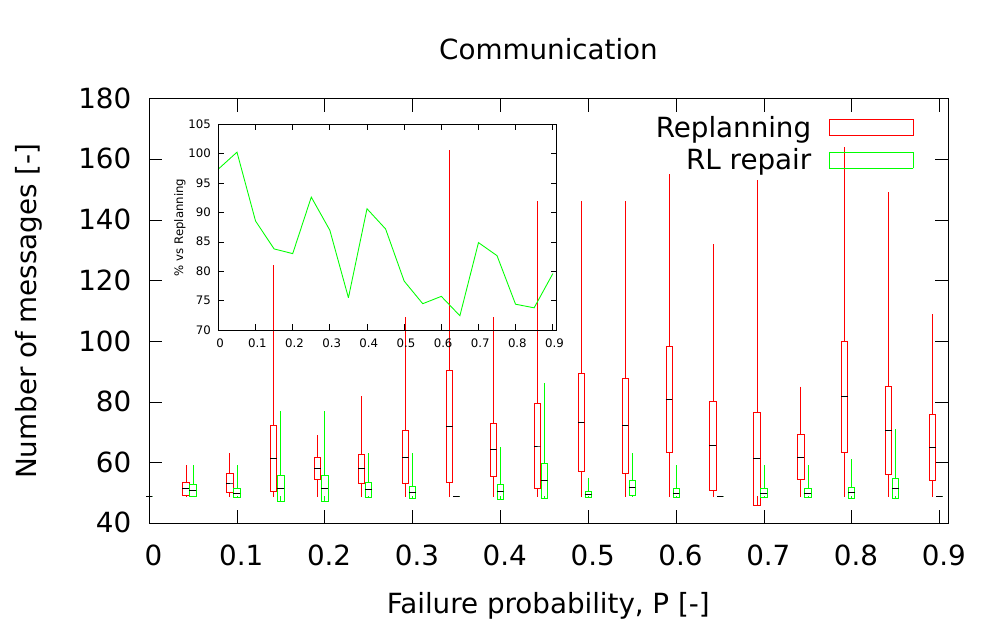}
\par\end{centering}

\caption{\label{fig:Experimental-log-sp-1}Experimental results for \textsc{logistics} domain with 3 agents and state perturbations with $c=1$.}
\end{figure}

The first batch of experiments directly targets validation of Hypothesis~\ref{hyp:repair-lower-overhead}: \emph{multi-agent plan repair is expected to generate lower communication overhead in tightly coordinated domains}. We used \textsc{logistics} as a tightly coordinated domain and dynamics of the simulated environment modeled as action failures. Figure~\ref{fig:Experimental-results-for-1} depicts the results of the experiment. The communication overhead generated by the \textsf{Back-on-Track-Repair} algorithm is on average only 59\% (36\% at best) of that generated by the replanning approach. Furthermore, the \textsf{Repeated-Lazy-Repair} algorithm performed even better and on average produced only 43\% (11\% at best) of the communication overhead generated by the replanning algorithm. In result, the experiments \emph{strongly support} our hypothesis. 

Additionally, the overall time spent in the planning phase (used by the \textsf{MA-Plan} algorithm) by the plan repair algorithms was 54\% (34\% at best) and 51\% (12\% at best) for \textsf{Back-on-Track-Repair} and \textsf{Repeated-Lazy-Repair} respectively. The execution length was lower in comparison the replanning approach as well being in average 96\% (72\% at best, 130\% at worst) by \textsf{Back-on-Track-Repair} and lower being 81\% (34\% at best, 132\% at worst) for \textsf{Repeated-Lazy-Repair}.

The second batch of experiments focused on boundaries of validity of the positive result presented above. In particular, we validated the condition on the coordination tightness and feasibility of failures. The auxiliary hypothesis we validated states: \emph{with decreasing coordination tightness of the domain, the communication efficiency gains of repairing techniques should decrease. For loosely coordinated domains the communication efficiency of plan repair should be on-par} \emph{with that of the replanning approach}.

To validate the auxiliary hypothesis we ran experiments with the \textsc{rovers} as a loosely coordinated domain. Figure~\ref{fig:Experimental-log-sp-1}~(top) presents the results supporting the hypothesis.

The third batch of experiments targeted the perturbation magnitude of the plan failures. The second auxiliary hypothesis we validated states: \emph{communication efficiency gain of plan repairing in contrast to replanning should decrease as the difference between a nominal and related failed state increases}. The underlying intuition is that, in the case the dynamic environment generates only relatively small state perturbations and the failed states are ``not far'' from the actual state, the plan repair should perform relatively well. On the other hand, if the state essentially ``teleports'' the agents to completely different states, replanning tends to generate more efficient solutions than plan repair.

To answer this hypothesis, we have prepared another \textsc{logistics} experiment employing state perturbations as the model of the environment dynamics. Figure~\ref{fig:Experimental-log-sp-1}~(bottom) depicts results of the experiment for $c=1$. The perturbed state for $c=1$ is produced by removing one term from the actual state and adding another one. As the chart shows, under random perturbations the plan repairing technique lost its improvement against replanning. For stronger perturbations with $c=2,3,4$, the ratio between plan repairing and replanning remained on average the same. The trend of the absolute numbers of messages, planning time and execution length was slightly decreasing, as the probability of opportunistic effects increased.

Finally, we conducted a series of experiments with a non-co\-or\-di\-na\-ted \textsc{sattelites} domain. The results depicted in Figure~\ref{fig:Experimental-log-sp-1} show the anticipated lower plan repair communication efficiency in contrast to replanning.

\section{Final remarks}

\label{sec:Discussion}

In the presented paper, we formally introduced the problem of multi-agent plan repair, proposed three algorithms for solving it and experimentally validated the hypothesis stating that under certain conditions, multi-agent plan repair approach tends to be more efficient in terms of the communication overhead it generates in comparison to the replanning approach. Our results well support the core hypothesis of the paper and we additionally performed a series of experiments validating its boundary conditions.

The line of research underlying this paper well correlates with recent works on classical single-agent planning sub-domains, such as partial ordered plan monitoring and repairing~\citep{Muise_McIlraith_Beck_2011}, conformant and contingency planning, plan re-use and plan adaptation. Environment dynamics is also handled by approaches based on Markov decision processes. The main difference to our approach is that the state perturbations utilized in our experiments have \emph{a priori }unknown probabilities. Our own recent approach to the problem of multi-agent plan repair in~\citep{conf/DMPOUW2011/KomendaNovak2011} can be seen only as a precursor to the formal and rigorous treatment of the problem in this paper. Therein, we described the first steps towards formal treatment of the problem, as well as proposed two specific incomplete algorithms for solving the problem, very distinct from the ones presented here.

There are several open challenges resulting from the presented work. Firstly, the multi-agent planning framework (\noun{MA-Strips}) is not expressive enough to describe certain aspects of concurrent actions and should be extended to this end, what, we suspect, will also influence the multi-agent planning complexity analysis. In particular, there is no way to account for joint actions which have effects strictly different than the unity of the individual actions involved. Another issue is that there is no way to enforce or forbid concurrent execution of certain individual actions. Secondly, the framework is not able to describe concurrent resource consumption, which is not an issue in single-agent \noun{Strips}~\citep{Fikes1971} planning, but in the multi-agent extension two individual concurrently executed actions might ``consume'' the same precondition, even though it is undesirable in the domain. Thirdly, there is a need for more efficient implementations of multi-agent planners with more features as the gap between the state-of-the-art classical planners and multi-agent planners is enormous. Fourthly, there is a lack of standardized planning benchmarks for multi-agent planning, especially considering tightly coordinated planning problems. Such are needed to further evaluate the hypotheses presented in this paper. Finally, we leave out the work towards resolving the validity of Hypothesis~\ref{hyp:repair-lower-comm-complexity} aiming at investigation of complexity issues of multi-agent plan repair to future work.

\end{document}